\documentclass{article}
\usepackage{arxiv}

\usepackage[utf8]{inputenc} 
\usepackage[T1]{fontenc}    
\usepackage{hyperref}       
\usepackage{url}            
\usepackage{booktabs}       
\usepackage{amsfonts}       
\usepackage{nicefrac}       
\usepackage{microtype}      
\usepackage{lipsum}		
\usepackage{graphicx}
\usepackage{natbib}
\usepackage{doi}

\usepackage{fancyhdr}       
\usepackage{graphicx}       
\graphicspath{{figures/}}     

\date{}

\renewcommand{\headeright}{}
\renewcommand{\undertitle}{}

\renewcommand{\shorttitle}{Predicting Seriousness of Injury in a Traffic Accident:}

\hypersetup{
pdftitle={Predicting Seriousness of Injury in a Traffic Accident},
pdfsubject={cs.AI},
pdfauthor={Paschalis Lagias,
George D. Magoulas,
 Ylli Prifti,
Alessandro Provetti},
pdfkeywords={Class imbalance, Data imputation, Feature engineering Neural networks, Reinforcement learning, Q--learning, Traffic accidents},
}

\long\def\COMMENT#1\ENDCOMMENT{\message{(Commented text...)}\par}
\begin{document}
\title{Predicting Seriousness of Injury in a Traffic Accident: A New Imbalanced Dataset and Benchmark}
%

\author{Paschalis Lagias,
\href{https://orcid.org/0000-0003-1884-0772}{\includegraphics[scale=0.06]{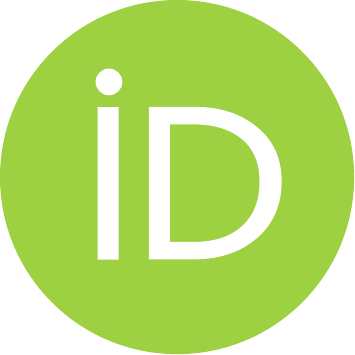}\hspace{1mm}George D. Magoulas},
\href{https://orcid.org/0000-0002-9323-875X}{\includegraphics[scale=0.06]{orcid.pdf}\hspace{1mm}Ylli Prifti} and
\href{https://orcid.org/0000-0001-9542-4110}{\includegraphics[scale=0.06]{orcid.pdf}\hspace{1mm}Alessandro Provetti}\\
Department of Computer Science and Information Systems\\
Birkbeck - University of London\\
London WC1 7HX, UK
}


\maketitle             
\begin{abstract}
The paper introduces a new dataset to assess the performance of machine learning algorithms in the prediction of the seriousness of injury in a traffic accident. 
The dataset is created by aggregating publicly available datasets from the UK Department for Transport, which are drastically imbalanced with missing attributes sometimes approaching 50\% of the overall data dimensionality. 
The paper presents the data analysis pipeline starting from the publicly available data of road traffic accidents and ending with predictors of possible injuries and their degree of severity. It addresses the huge incompleteness of public data with a MissForest model. 
The paper also introduces two baseline approaches to create injury predictors: a supervised artificial neural network and a reinforcement learning model. 
The dataset can potentially stimulate diverse aspects of machine learning research on imbalanced datasets and the two approaches can be used as baseline references when researchers test more advanced learning algorithms in this area.
\end{abstract}
%
%
\section{Introduction} \label{sec:introduction}
Nowadays detailed information about traffic accidents is becoming available for independent analysis.
Authorities that collect such data may release, along with traditional statistical aggregations, actual data points that are a rich source of information. 
Apart from time, location, number of vehicles involved and similar factual information, the data record often concerns subjective measures such as the severity of the accident, which is annotated by trained traffic police officers.

In the UK, the Department for Transport (DfT) aggregates and releases a dataset of reference with many details about each accident  recorded. 
While data are available, there is a huge imbalance in the information provided between many minor events, e.g., collisions in parking lots, and the-- fortunately less frequent-- major events that involve hospitalisation or worse. 

Several researchers have examined parts of the UK's DfT traffic accident data in order to answer a variety of research questions. 
Among them, when it comes to predicting accident severity, ~\cite{almohimeed-medium},~\cite{babic-descriptive},~\cite{haynes-data} and~\cite{ml-kumeda}, a central research question is: \emph{``in the scenario of a traffic accident with injuries, how severe is the injury going to be, based on available data on accident conditions, vehicle information etc.?''} 
Studies, like the ones cited above, have focused on analysing a specific year or period of traffic data and although they have considered accident severity in general, they did not focus on predicting the seriousness of injuries. 
This problem comes across as being very challenging because the DfT considers that severity of injury is a \textit{triage}, namely slight, serious, or fatal %
\footnote{See ``Instruction for the completion of accident reports'', Dept. for Transport (2005).}, 
which leads to a highly-imbalanced distribution of data that impacts the prediction accuracy, especially over minority classes (e.g. fatal accidents), of the methods tested.

Thus the paper considers the prediction of the seriousness of injury as an imbalanced multi--class classification problem. It extends previous work,~\cite{almohimeed-medium},~\cite{babic-descriptive},~\cite{haynes-data} and~\cite{ml-kumeda}, by applying a systematic data analysis and processing pipeline to combine data from disparate sources of the UK's DfT from years 2005--2018 in order to create a new larger dataset. 
The pipeline incorporates components for data imputation, based on domain knowledge and the predictive power of variants of Random Forests, and feature importance analysis components, which use categorical feature correlation, mutual feature information and $\chi^2$--tests, with more detailed description of each pipeline component to be presented in a later section.

Lastly, the paper proposes two evaluation approaches to create machine learning predictors using the new dataset. 
These could be used as baseline references when designing machine learning methods to predict the seriousness of injuries in the scenario of a traffic accident given certain accident conditions, such as involved vehicle information and some personals details (anonymised) of the potentially-injured person and so on. 

The rest of the paper is organised as follows. 
Section \ref{sec:related} presents relevant work, while Section \ref{sec:sources} describes the data sources that were used. 
Section \ref{sec:preparation} describes the components of the pipeline that were used to create the new dataset. 
The baseline models are presented in Section \ref{sec:archi}, and their evaluation is presented in Section \ref{sec:eval}. 
The paper ends with conclusions in Section \ref{sec:conclusions}. 

\section{Relevant Work}\label{sec:related}

UK traffic accident datasets are imbalanced with several missing attributes. Previous studies, ~\cite{almohimeed-medium}~\cite{babic-descriptive}~\cite{haynes-data}~\cite{ml-kumeda}, attempted to deal with the challenges in these data by limiting the dimensionality of the problem, focusing for example on data from a specific year or period, exploring the potential of specific subsets of attributes that were available across all data points considered, or by transforming the multi--class problem into a binary one. Although overall satisfactory accuracy was produced, all models experienced very low accuracy over minority classes. 

More relevant to this paper is the recent effort in~\cite{haynes-data}, where the authors used a variety of tools, such as different statistical methods and Machine Learning (ML) algorithms, in the search for the right ``mix'' of feature selection and ML algorithm that would provide good predictors for accident severity. 
The best results were achieved by Random Forests (RFs) running over an input of 14 different features, from, e.g., the age of the driver to the weather conditions. A RF achieved an overall accuracy of 85.08\% with 15.12\%, 22.03\% and 96.58\% correct prediction for the fatal, serious and slight class of injuries, respectively. However, the experiments used only c. 136k records from accidents reported in 2016, which is rather limited given the availability of data reported by the DfT%
\footnote{Please see \href{https://data.gov.uk/dataset/cb7ae6f0-4be6-4935-9277-47e5ce24a11f/road-safety-data}{https://data.gov.uk}}.

In comparison to that approach, and others cited above, this paper describes a pipeline that includes acquiring, data-cleaning, and inputting long-term accident data (2005--2018) to create a new large dataset for multi--class classification. This can potentially enable ML methods to pick up small fluctuations or relatively rare events (e.g. did not appear in 2015 or 2016), but can determine a non-trivial amount of accident cases, such as ice on the road that does appear only sporadically in the UK but certainly determines a spike in the number and gravity of accidents. Furthermore, the dataset is expanded horizontally by including many new features, in search for non-standard influences.

\section{Data Sources} \label{sec:sources}

The UK's Department for Transport publishes three datasets per year, uploaded in the {\textit{Road Safety Data}} web page of the \texttt{data.gov.uk} website:

\begin{itemize}
	\item \textbf{Accidents}, with variables related to accident conditions. 
	Each accident is identified by a unique accident ID, called ``Accident Index''.
	
	\item \textbf{Vehicles}, with variables related to vehicle characteristics, driver information and driver action before the accident. 
	Each vehicle is identified with a unique vehicle reference number, ``Vehicle Reference'', and linked with accidents dataset through an ``Accident Index''.
	
	\item \textbf{Casualties}: information about injured individuals, linking an injury with accidents and vehicles through ``Accident Index'' and ``Vehicle Reference''.
\end{itemize}


The paper exploits  DfT data from 2005 to 2018, with data from 2019 used for testing. 
Table \ref{tab:class-dist} shows data distribution in the new aggregated dataset, highlighting the drastic imbalance among the target classes data.

\begin{table}[h]
    \centering
    \begin{tabular}{|c|r|r|}
    \hline
    slight  & 2,539,715 & 87.10\% \\
    \hline
    serious &   345,997 & 11.87\%\\
    \hline
    fatal  &    30,171 &  1.03\%\\
    \hline
    \end{tabular}
    \caption{Distribution of casualty severity in the new aggregated DfT data.}
    \label{tab:class-dist}
\end{table}

\section{Creating the Dataset} \label{sec:preparation}

The first phase in creating the new dataset involved accessing and merging data from disparate sources into a single dataset. 
Next, the work dealt with missing values, running missing value imputation, whenever possible, and assessing the potential importance of each feature for the classification phase. 

Imputation was based on domain knowledge and the predictive power of Random Forests. 
With regards to assessing the potential importance of each feature, which is relevant for machine learning classifiers, various techniques were used. 
Since most of the variables are nominal,
$\chi^2$--tests, Cramer's mutual information, and Theil's U coefficients were used \cite{zychlinski-tds}. 
For correlation of numerical features with nominal ones, the ANOVA test and correlation ratios \cite{zychlinski-tds} were computed. 
Once feature importance analysis and imputation are completed, data can be transformed to an input suitable for classification methods. 
The details of the data analysis pipeline are presented below.

\subsection{Dataset merging} \label{sec:merge}
A fragment of the variables for the three datasets (Accidents, Vehicles, Casualties), starting from 2005 and up to 2018, is shown in Table \ref{tab:vars}, where the target variable, ``Casualty Severity'',  is shown in italics.
The datasets were merged by using Accident Index and Vehicle Reference as ``foreign keys'': using Vehicle Reference, each casualty was matched with a vehicle; pedestrians were matched with vehicles that caused their injury.
Next, the output records of the above join were matched with accidents using Accident Index. 
The final output consists of 2,915,883 data points and 66 variables/features in total (cf. with  Table \ref{tab:class-dist}).

\begin{table}[htb]
    \centering
    \begin{tabular}{|l|l|l|}
    \hline
    \textbf{Accidents}  & \textbf{Vehicles} & \textbf{Casualties} \\
    \hline
    Accident Index &   Accident Index & Accident Index \\
    \hline
    Location Easting OSGR  &    Vehicle Reference &  Vehicle Reference \\
    \hline
    Location Northing OSGR  &    Vehicle Type &  Casualty Reference \\
    \hline
    Longitude  &    Towing and Articulation &  Casualty Class \\
    \hline
    Latitude  &    Vehicle Manoeuvre &  Sex of Casualty \\
    \hline
    Police Force  &    Vehicle Location-Restricted Lane &  Age of Casualty \\
    \hline
    Accident Severity  &    Junction Location &  Age Band of Casualty \\
    \hline
    Number of Vehicles  &    Skidding and Overturning &  \textbf{\textit{Casualty Severity}} \\
    \hline
    \dots & \dots & \dots \\
    \hline
    \end{tabular}
    \caption{Dataset variables}
    \label{tab:vars}
\end{table}

\subsection{Dealing with missing values} \label{sec:miss-check}

Once merged, the DfT data turned out to be of rather poor quality with only 411,158 data points containing all necessary information, i.e., 86\% of the available data points had at least one missing variable value.
%
\COMMENT
\begin{table}[htb]
    \centering
    \begin{tabular}{|r|l|}
    \hline
    \textbf{Total rows}  & 2,915,883 \\
    \hline
    \textbf{Total columns} & 66 \\
    \hline
    \textbf{Total full rows}  &  411,158 \\
    \hline
    \textbf{Total incomplete rows}  &  2,504,725 \\
    \hline
    \textbf{\% of incomplete rows}  &  85.90 \% \\
    \hline
    \end{tabular}
    \caption{Merged dataset dimensions and missing values}
    \label{tab:dimensions}
\end{table}
\ENDCOMMENT{}
Most of the variables in the dataset are categorical with integer encoding, and missing values are represented with -1. 
In addition, some variables include encoding for ``unknown'' as a separate category, e.g. like the feature Weather Conditions. 
Initially, the following features were dropped:

\begin{itemize}
	\item \textbf{Accident Index, Vehicle Ref. and Casualty Ref.}: administrative references that are not useful for classification.
	
	\item \textbf{Age Band of Driver and Age Band of Casualty}: data already include ages for casualties and drivers.
	
	\item \textbf{Latitude, Longitude and LSOA of Accident Location}: data already include Eastings and Northings.
	
	\item \textbf{Did Police attend the Scene of the Accident, Accident Severity and Number of Casualties}: post-accident information is not included in the research.
	
	\item \textbf{Road Maintenance Worker, Journey Purpose of Driver and Engine Capacity}: these variables have an excessive number of either missing, ``unknown'' or ``not applicable'' values. 
	Regarding Engine Capacity, many of the available values were found to be  inconsistent with the Vehicle Types.
\end{itemize}

Lastly, for Urban or Rural Area, Sex of Casualty, Sex of Driver, Pedestrian Location, Pedestrian Movement and Light Conditions, all values for ``unknown'' were re-encoded as missing values with $-1$.
\subsection{Missing-value imputations based on domain knowledge}\label{sec:missing}

Domain knowledge was derived from available DfT documentation about accident data gathering and relevant guidelines. That was further informed by critical analysis and reasonable assumptions based on the known variables, and used for manual imputation of the missing values as follows:

\begin{itemize}
	\item Missing values for Car Passenger, which were related to Casualty Type of buses and vans, were replaced with ``Not car passenger'' value.
	
	\item It was assumed that bicycles, motorcycles and mobility scooters have no Towing or Articulation.
	
	\item Bicycles, horses, motorcycles and trams cannot be classified as left or right hand drive vehicles.
	A new category was created for ``unknown''.
	
	\item Many missing values for Junction Location, Junction Detail, Junction Control and 2nd Road Class were corrected, as they referred to accidents that did not occurred near a junction.
	
	\item Some missing values for Age of Driver, Age of Casualty, Casualty Home Area Type and Driver Home Area Type were corrected by checking samples where the casualty was the driver.
	
	\item All missing values for light conditions were set to ``Daylight'' after checking Time value of the accident.
\end{itemize}

Lastly, data points where Eastings, Northings or Time values were missing were dropped from the data. 
The result of this phase was a record with 53 features and 2,915,387 data points; still 1,471,895 data points, or 50.49\% of the total, had one or more missing values.


\COMMENT
\begin{table}[htb]
    \centering
    \begin{tabular}{|c|r|r|}
    \hline
    \textbf{rows}  & 2,915,387 &\\
    \hline
    \textbf{full rows}  &  1,443,492 & 49.51\% \\
    \hline
    \textbf{incomplete r.}  &  1,471,895 & 50.49\% \\
    \hline
    \end{tabular}
    \caption{Merged dataset dimensions and missing values}
    \label{tab:datasize}
\end{table}
\ENDCOMMENT

\subsection{Time processing, feature correlation and feature importance}\label{sec:feat-corr}

The next step in the pipeline was to inspect the numerical variables of the produced dataset, assess the importance of features and measure possible correlations and collinearities among variables.

Variables Date and Time were discarded after they were converted into new variables Hour, Month and Year. Next, the following features were treated as numerical: Number of Vehicles, Speed Limit, Age of Casualty, Age of Vehicle, Age of Driver, Location Easting OSGR and Location Northing OSGR. All remaining variables were treated as nominal and represented by discrete values without any form of intrinsic ranking. 
Casualty Severity, the target variable, is also nominal.

Due to the mixture of nominal and numerical variables in the data, various feature selection and correlation metrics were used (the threshold for considering two features as highly correlated was set to 0.7), depending on the combination  of examined features:

\begin{itemize}
	\item Pearson's correlation was calculated for all pairs of numerical variables.
	
	\item A $\chi^2$--squared statistic and mutual information between each categorical variable and Casualty Severity were used to assess the importance of each categorical variables wrt. the target variable.
	
	\item the ANOVA F--statistic was calculated to assess the importance of each numerical variable with respect to Casualty Severity.
	
	\item Correlation Ratio coefficient for each pair of categorical--numerical variables was employed to check possible correlation in input variables, and confirm the importance of numerical variables wrt. Casualty Severity.
	
	\item Cramer's V and Theil's U correlation were calculated for each pair of categorical variables to detect any correlation among input variables.
	Unlike the rest of correlation coefficients, Theil's U is anti-symmetrical and is based on mutual information (entropy) between two variables.

\end{itemize}

After assessing features importance, {Casualty Type, Vehicle Type} and {Vehicle Manoeuvre} were found to be the most important categorical variables, while {Number of Vehicles} and {Speed Limit} were the most important numerical ones.
On the opposite end, {Carriageway Hazards, Was Vehicle Left Hand Drive} and {Pedestrian Crossing-Human Control} were the least important categorical variables; {Eastings} and {Northings} were the least important numerical ones. 

Lastly, analysis of feature importance and computation of correlation revealed that:

\begin{itemize}

	\item Variables {Police Force, Local Authority (District), Local Authority (Highway), Eastings} and {Northings} are highly correlated. 
	Only {Local Authority (District)} was kept as it was found important wrt. Casualty Severity.

	\item {Casualty Type} is highly correlated with {Casualty Class} and {Vehicle Type}, and it was dropped.
	
	\item Variables {Casualty Home Area Type} and {Driver Home Area Type} are highly correlated with each other, so the former was dropped.
	
	\item Variables {1st Road Number} and {2nd Road Number} were dropped due to their high cardinality.
\end{itemize}

\subsection{Imputation with MissForest}
The MissForest algorithm \cite{stekhoven-missforest} was used to obtain missing--value imputation.
MissForest is a type of Random Forest algorithm suitable for handling high--dimensional datasets with mixed data types (categorical and numerical), which is exactly our case.
Given the size of the dataset and taking into consideration memory requirements, 
the procedure was applied iteratively:

\begin{enumerate}
	\item {Speed Limit} and {Weather Conditions} were initially imputed, based on non-missing variables, to increase the number of complete rows.
	
	\item The remaining variables were divided in groups based on topic: Junctions and road classes, Pedestrians, Vehicle--site interaction, Drivers--vehicles, Age of Vehicle and Driver IMD Decile.

	\item For each imputed variable, 100 trees were grown.
\end{enumerate}

MissForest imputation produces a new dataset with 2,915,387 data points and 49 features. Note that {Casualty Type}, {Casualty Home Area Type, Police Force} and {Local Authority (Highway)} were excluded from the imputation process.
After missing--value imputation was completed, the remaining ``unimportant'' and highly--correlated features listed in Section \ref{sec:feat-corr} were removed.
\section{Baseline Models} \label{sec:archi}

In this section two approaches are described to create predictors of the seriousness of injury. Creating optimal, or fine--tuned models, is out--of--scope for this article. Instead the aim is to provide a point of reference for researchers to further explore this dataset using machine learning methods. First, supervised learning using a neural network classifier is considered, and then a non--traditional form of learning by reinforcement using Deep--Q Network is explored. 

\subsection{A supervised learning model}\label{sec:super}

A small number of preliminary experiments were conducted to identify an architecture that performs reasonably well, given the imbalanced nature of the dataset, but no serious attempt was made to optimise model or training algorithm hyperparameters. The outcome was a densely--connected artificial neural network (ANN) implemented in Keras with TensorFlow backend:

\begin{itemize}
	\item \textbf{Hidden layers and neurons:} two or three hidden layers were used, with some of the best results presented in the next section.
	
	\item \textbf{Output layer:} three nodes representing the three classes used by the DfT.
	
	\item \textbf{Activation functions:} ReLU activations for hidden--layer nodes and Softmax activations for output nones.
	
	\item \textbf{Optimiser:} the Adam optimiser was used in all experiments.
	
	\item \textbf{Batch size:} 512 showed better behaviour than 128 or 256.
	
	\item \textbf{Early stopping:} experimented with 5 and 20 epochs of early stopping.
	
	\item \textbf{Weight initialisation:} both Glorot and He uniform were tested. 
	
	\item \textbf{Loss function:} the sparse categorical cross entropy was adopted.

	\item \textbf{Class weights:} since the distribution of Casualty Severity is heavily imbalanced, different class weights were tried. 
	The vector of class weights was initially computed as $\frac{|\mathrm{samples}|}{|\mathrm{classes}| * |\mathrm{frequencies}|}$. 
	
\COMMENT
	\begin{equation}
	   \mathrm{weights} = \frac{|\mathrm{samples}|}{|\mathrm{classes}| * |\mathrm{frequencies}|}
	\end{equation}
\ENDCOMMENT

\end{itemize}

\subsection{A reinforcement learning model}\label{sec:rl}

Reinforcement Learning (RL) is not traditionally applied to classification problems, but recent work has shown that it is possible to formulate a classification task as a sequential decision-making problem and solve it with a deep Q--learning network~\cite{lin-deep19}. Moreover, empirical studies demonstrated that this approach can reach strong performance, outperforming other imbalance classification methods, especially when there is high class imbalance~\cite{lin-deep19}. Our RL model followed this approach and was implemented in OpenAI Gym:

\begin{enumerate}
	\item \textbf{Environment:} This was defined as the dataset itself, including the following  attributes:
	
	\begin{itemize}
	    \item \textbf{Observation space:} the size of a data sample.
	    
	    \item \textbf{Action space:} taking an action as equivalent of a class prediction, there are three possible actions, one per prediction: slight, serious or fatal.
	    
	    \item \textbf{Step counter:} an integer to track the number of steps the agent has taken in the environment within the same episode.
	    
	    \item \textbf{Weighted action rewards:} a different reward can be earned for correct classification of each class data sample.
	    
	    \item \textbf{Reset function:} it resets the environment at the end of the episode-- shuffling data; resetting step counter; retrieving the first training sample.
	    
	    \item \textbf{Step:} a function that makes a Casualty Severity prediction, collects reward and checks if the episode is done. 
	    If so, it moves to the next training sample; otherwise, the environment is reset.
	\end{itemize}
	
	\item \textbf{Episode:} it starts when the first training sample is read, and it ends when all training samples are classified, or a minority class sample is misclassified.
	
	\item \textbf{Reward function:} the recommendations of~\cite{lin-deep19} were followed:
	
	{\small
	\begin{equation}\label{reward}
	    reward = \left\{
	    \begin{array}{cccccc}
	         1 & if\; label & = & fatal\; and\; prediction & = & fatal\\
	         -1 & if\; label & = & fatal\; and\; prediction & \neq & fatal\\
	         r_1 & if\; label & = & serious\; and\; prediction & = & serious\\
	         -r_1 & if\; label & = & serious\; and\; prediction & \neq & serious\\
	         r_2 & if\; label & = & slight\; and\; prediction & = & slight\\
	         -r_2 & if\; label & = & slight\; and\; prediction & \neq & serious,\\
	    \end{array}\right.
	\end{equation}
	} 
	
	\noindent
	where $r_1$ is the ratio of \textit{Serious injury samples} to \textit{Slight injury samples}, and $r_2$ is the ratio of \textit{Fatal injury samples} to \textit{Slight injury samples.}
	
	\item \textbf{Memory:} a facility to save transitions in the environment and sample batches of saved transitions for Q--Network training.
	
	\item \textbf{Transition:} it includes the following information:
	
	\begin{itemize}
	    \item \textbf{Current state:} the current training sample.
	
	    \item \textbf{New state:} the next training sample.
	
	    \item \textbf{Action taken:} predicted severity.
	
	    \item \textbf{Reward:} value earned/lost based on prediction and sample actual label.
	
	    \item \textbf{Episode done flag:} a boolean that indicates episode's completion.
	\end{itemize}
	
	\item \textbf{Agent:} an agent has memory, as described above, and holds the following functionality and attributes:
	
	\begin{itemize}
	
	    \item \textbf{Evaluation and target networks:} it uses one network for training and a second target network, which is updated periodically after a fixed number of steps. This avoids overestimation of Q values and enhances training stability \cite{dql-dmind}.
	
	    \item \textbf{Action space:} the set of available actions (class predictions).
	
	    \item \textbf{Hyperparameters:} they relate to training and reward collection, e.g. $\epsilon$ for $\epsilon$--greedy policy, $\gamma$ for reward discount, batch size to sample from Agent's memory, $N$ number of episodes before updating target network and Q--Network optimiser with learning rate.
	
	    \item \textbf{Save:} a facility to save a transition in Memory and training/target network weights, and the entire memory space.
	
	    \item \textbf{Load:} a facility to load saved network weights and Memory.
	
	    \item \textbf{Training:} it updates the training/target network weights.
	
	    \item \textbf{Action selection:} predicting Casualty Severity  for a training sample.
	\end{itemize}
\end{enumerate}

The steps of a full episode within the environment of traffic accidents are:

\begin{enumerate}
    \item The environment is reset.

    \item The agent checks the first training sample and predicts Casualty Severity using the $\epsilon$--greedy policy.
    
    \item A reward, a new training sample and an episode done flag are returned.
    
    \item Agent prediction is compared against actual label and the training network weights are updated by  back-propagation with Stochastic Gradient Descent.
    
    \item If a minority sample (serious, fatal) is classified incorrectly, the episode ends and the environment is reset.
    Otherwise, the agent takes a new step.
\end{enumerate}

Training and target neural networks use the architecture described in Section~\ref{sec:super}.
However, no Softmax activation is needed, since the RL approach is based on collecting maximum reward from Q values. 

\section{Evaluation} \label{sec:eval}

The following experiments could be used as a baseline when optimising similar models or designing more sophisticated approaches. 
Data and models are available at: \url{https://ale66.github.io/traffic-accident-gravity-predictor/}. 
In this context, different models were trained and tested (cf. with Section~\ref{sec:archi}) without hyperparameter optimisation or fine--tuning. In training and validation, 2005--2018 accident data (cf. with Table~\ref{tab:class-dist}) were used with a 75\%--25\% split. 
Different test sets were used, as described below, and the following metrics: 

\begin{itemize}
    \item \textbf{Overall classification accuracy:} correctly classified test samples over total number of test samples
    
    \item \textbf{Class accuracy:} performance in testing on each single class.
\end{itemize}

\subsection{Supervised ANN experiments} \label{sec:sl-res}

Experiments with ANNs were run using different versions of the dataset.

\paragraph{Experiment 1:} only full records, c. 411k data points, were used for training and validation in this experiment, i.e. there was no imputation or resampling to treat imbalance. 
Testing used 2019 data (c. 153k data points).
Highest accuracy for fatal injury in testing, 53\%, was achieved with an ANN of three hidden layers with 1000 neurons each and class weights 17.5, 2.44 and 0.69 for the fatal, serious and slight classes, respectively.
The highest accuracy for the serious--injury class, 66\%, was achieved with an ANN of two hidden layers with 2000 and 500 neurons respectively, and class weights of 31.83, 3.04 and 0.38, for the fatal, serious and slight classes, respectively. 
The highest overall classification accuracy, 77\%,  and best accuracy per class on average, 56\% (44\% fatal; 43\% serious; 82\% slight), were achieved with an ANN of three hidden layers with 1200 neurons each and class weights of 19.5, 3.44 and 0.69 for the fatal, serious and slight classes, respectively.

\paragraph{Experiment 2:} as above, the full records (c. 411k data points) were used but this time Synthetic Minority Over--sampling Technique (SMOTE),~\cite{smote}, was applied to treat class imbalance. 
The default value of three neighbours was used for generation of synthetic samples and the output was a new dataset with an equal amount of observations per class.
The best test results were achieved with an ANN of three hidden layers with 1200 neurons each: 24\% for fatal, 63\% for serious, 67\% for slight. 
One of these models also exhibited the best available accuracy for the fatal--injury class (about 25\%) with class weights 1.23, 1.07 and 0.89 for the three classes respectively. 
For the serious--injury class, the best available model achieved an accuracy of 63\% with class weights 1, 0.85 and 0.5, for the three classes respectively. 
Although tests failed to show clear advantage when SMOTE training data are used, fine--tuning deserves some consideration. 

\paragraph{Experiment 3:} this experiment focused on the larger dataset (cf. with Table~\ref{tab:class-dist}) with imputation, and testing was based on c. 153k data points from 2019. 
Again, the best results were achieved with an ANN of three hidden layers with 1200 neurons each: 45\% for fatal, 57\% for serious, and 66\% for slight.
Best available accuracy for the fatal--injury class was 48\% with weights 19.5, 3.44 and 0.69 for the three injury classes respectively.
For the serious--injury class, best available accuracy was 64\%, with class weights 32.1, 2.81 and 0.38, for the three classes respectively. 
In comparison, best available results for logistic regression (the stochastic incremental gradient method SAGA and an L2 penalty was used) reached 67\% for fatal, 43\% for serious, and 63\% for slight, indicating that further tuning of the ANN model is needed.
Simulations were also run using SMOTE generated data but, as in Experiment 2, test results did not show clear benefits for the minority classes. 

\COMMENT
\begin{table}[htb]
    \centering
    \begin{tabular}{|r|r|r|r|r|}
    \hline
    \textbf{Weights-Metrics}  & \textbf{Best Fatal} & \textbf{Best Severe} & \textbf{Best Val. Acc.} & \textbf{Best Avg. Acc.} \\
    \hline
    \textbf{Fatal Weight}  & 17.500 & 31.830 & N/A & 19.500 \\
    \hline
    \textbf{Severe Weight} & 2.440 & 3.040 & N/A & 3.440 \\
    \hline
    \textbf{Light Weight} & 0.690 & 0.380 & N/A & 0.690 \\
    \hline
    \textbf{Fatal Accuracy\%} & 52.315 & 26.305 & 2.241 & 44.430 \\
    \hline
    \textbf{Severe Accuracy\%} & 23.483 & 65.816 & 6.249 & 42.596 \\
    \hline
    \textbf{Light Accuracy\%} & 88.716 & 68.026 & 99.398 & 82.119 \\
    \hline
    \textbf{Val. Accuracy\%} & 81.150 & 67.340 & 88.140 & 77.370 \\
    \hline
    \textbf{Avg. Class Acc. \%} & 53.111 & 53.382 & 36.023 & 56.383 \\
    \hline

    \end{tabular}
    \caption{Results with dataset 1}
    \label{tab:sl-results1}
\end{table}

\begin{table}[htb]
    \centering
    \begin{tabular}{|r|r|r|r|r|}
    \hline
    \textbf{Weights-Metrics}  & \textbf{Best Fatal} & \textbf{Best Severe} & \textbf{Best Val. Acc.} & \textbf{Best Avg. Acc.} \\
    \hline
    \textbf{Fatal Weight}  & 1.230 & 1.000 & N/A & 1.000 \\
    \hline
    \textbf{Severe Weight} & 1.070 & 0.850 & N/A & 0.850 \\
    \hline
    \textbf{Light Weight} & 0.890 & 0.500 & N/A & 0.500 \\
    \hline
    \textbf{Fatal Accuracy\%} & 25.296 & 24.828 & 23.719 & 24.828 \\
    \hline
    \textbf{Severe Accuracy\%} & 55.907 & 63.456 & 46.151 & 63.456 \\
    \hline
    \textbf{Light Accuracy\%} & 73.198 & 67.016 & 80.026 & 67.016 \\
    \hline
    \textbf{Val. Accuracy\%} & 70.790 & 66.180 & 75.700 & 66.180 \\
    \hline
    \textbf{Av. Class Acc. \%} & 51.467 & 51.767 & 49.965 & 51.767 \\
    \hline

    \end{tabular}
    \caption{Results with dataset 2}
    \label{tab:sl-results2}
\end{table}

\begin{table}[htb]
    \centering
    \begin{tabular}{|r|r|r|r|r|}
    \hline
    \textbf{Weights-Metrics}  & \textbf{Best Fatal} & \textbf{Best Severe} & \textbf{Best Val. Acc.} & \textbf{Best Av. Acc.} \\
    \hline
    \textbf{Fatal Weight}  & 19.500 & 32.105 & 19.500 & 32.105 \\
    \hline
    \textbf{Severe Weight} & 3.440 & 2.814 & 3.440 & 2.814 \\
    \hline
    \textbf{Light Weight} & 0.690 & 0.383 & 0.690 & 0.383 \\
    \hline
    \textbf{Fatal Accuracy\%} & 48.098 & 26.768 & 48.098 & 44.547 \\
    \hline
    \textbf{Severe Accuracy\%} & 36.982 & 64.497 & 36.982 & 57.536 \\
    \hline
    \textbf{Light Accuracy\%} & 82.221 & 65.579 & 82.221 & 66.174 \\
    \hline
    \textbf{Val. Accuracy\%} & 76.480 & 65.050 & 76.480 & 64.920 \\
    \hline
    \textbf{Av. Class Acc. \%} & 55.767 & 52.281 & 55.767 & 56.086 \\
    \hline

    \end{tabular}
    \caption{Results with dataset 3}
    \label{tab:sl-results3}
\end{table}
\ENDCOMMENT

\subsection{Reinforcement learning experiments} \label{rl-res}

Q--learning proved to be demanding computationally, although saving and loading memory and network weights may ease some of the burden of training the evaluation and the target networks for thousands of episodes. 
Different variants of the reward function, Eq.~(\ref{reward}), were tried, e.g. different reward ratios, slightly increasing the reward for successful fatal class predictions, reducing even more the reward for predicting correctly light injuries, with no clear benefit. 
Memory size was set to 1,000,000, $\gamma$ was 0.1, initial $\epsilon$  was 1.0 and final $\epsilon$ was 0.01 after all decrements.

To alleviate computational demands, only the full records, c. 411k data points, were used, keeping 75\% for training and 25\% for testing, without imputation or resampling. 
In testing, a Softmax activation was added to the output layer of the target network to generate an injury--class prediction.

As per Section~\ref{sec:rl}, ANNs with three hidden layers of 1200 neurons each were used. 
Best available accuracy per class in testing was 29\% for fatal, 49\% for serious, and 58\% for slight, which was achieved after 5800 training episodes.
Best available accuracy in testing for the fatal--injury class was 37\% after training for 3400 episodes. 
For the serious--injury class, best available accuracy in testing was 69\% by a model trained across 4700 episodes. Increasing the number of training episodes to several thousands has led to overestimation. 
Clearly that is an issue that deserves further investigation as it has been encountered in RL applications before and various strategies have been proposed,  e.g. tuning the rewards, maximising representation diversity or some form of regularisation, which may improve the RL model. 


\COMMENT
\begin{table}[htb]
    \centering
    \begin{tabular}{|r|c|c|c|c|}
    \hline
    \textbf{Parameters-Metrics}  & \textbf{Best Fatal} & \textbf{Best Severe} & \textbf{Best Val. Acc.} & \textbf{Best Av. Acc.} \\
    \hline
    \textbf{Training episodes}  & 3400 & 4700 & 2200 & 5800 \\
    \hline
    \textbf{Fatal injury reward}  & 1.0 & 1.0 & 1.0 & 1.2 \\
    \hline
    \textbf{Severe injury reward} & $S/L$ & $S/L$ & $S/L$ & $S/L$ \\
    \hline
    \textbf{Light injury reward} & $F/L*0.1$ & $F/L*0.1$ & $F/L$ & $F/L$ \\
    \hline
    \textbf{Fatal Accuracy\%} & 37.217 & 10.296 & 15.172 & 29.187 \\
    \hline
    \textbf{Severe Accuracy\%} & 54.493 & 69.115 & 20.738 & 49.747 \\
    \hline
    \textbf{Light Accuracy\%} & 18.480 & 45.087 & 89.020 & 58.122 \\
    \hline
    \textbf{Val. Accuracy\%} & 22.640 & 47.360 & 80.730 & 56.890 \\
    \hline
    \textbf{Av. Class Acc. \%} & 36.730 & 41.499 & 41.643 & 45.685 \\
    \hline

    \end{tabular}
    \caption{Reinforcement learning: best results}
    \label{tab:rl-results}
\end{table}
\ENDCOMMENT

\section{Conclusions} \label{sec:conclusions}
While several studies have sought to deploy ML to process public traffic accident data, to the best of our knowledge this is the first attempt to create a clean 2005--2018 dataset for predicting the seriousness of personal injuries. 
There are of course several alternatives that one can explore with respect to improving the data quality and the predictive ability of ML methods. 
Experiments with the two base models demonstrated that obtaining good accuracy on the minority classes without compromising performance on the majority class is very challenging, and perhaps requires applying more sophisticated approaches. 

Although systematic comparison and fine--tuning were out--of--scope for this paper, experiments highlighted the potential of supervised learning. 
Avenues for further investigation naturally include hyperparameter tuning and model optimisation.

\bibliographystyle{unsrtnat}
\bibliography{traffic-biblio}

\begin{thebibliography}{9}
\providecommand{\natexlab}[1]{#1}
\providecommand{\url}[1]{\texttt{#1}}
\expandafter\ifx\csname urlstyle\endcsname\relax
  \providecommand{\doi}[1]{doi: #1}\else
  \providecommand{\doi}{doi: \begingroup \urlstyle{rm}\Url}\fi

\bibitem[Almohimeed(2019)]{almohimeed-medium}
R.~Almohimeed.
\newblock {UK} traffic accidents -- data analysis (10+years), 2019.
\newblock URL \url{https://medium.com/@rawanme/}.

\bibitem[{Babič} and {Zuskáčová}(2016)]{babic-descriptive}
Frantisek {Babič} and Karin {Zuskáčová}.
\newblock Descriptive and predictive mining on road accidents data.
\newblock In \emph{IEEE 14th Int Symp on Applied Machine Intelligence and
  Informatics (SAMI)}, pages 87--92, 01 2016.
\newblock \doi{10.1109/SAMI.2016.7422987}.

\bibitem[Haynes et~al.(2019)Haynes, Estin, Lazarevski, Soosay, and
  Kor]{haynes-data}
Steven Haynes, Prudencia~Charles Estin, Sanela Lazarevski, Mekala Soosay, and
  Ah{-}Lian Kor.
\newblock Data analytics: Factors of traffic accidents in the {UK}.
\newblock In \emph{10th Int Conf on Dependable Systems, Services and
  Technologies, Leeds, United Kingdom, June 5-7, 2019}, pages 120--126. {IEEE},
  2019.
\newblock \doi{10.1109/DESSERT.2019.8770021}.

\bibitem[Kumeda et~al.(2019)Kumeda, Zhang, Zhou, Hussain, Almasri, and
  Assefa]{ml-kumeda}
Bulbula Kumeda, Fengli Zhang, Fan Zhou, Sadiq Hussain, Ammar Almasri, and
  Maregu Assefa.
\newblock Classification of road traffic accident data using machine learning
  algorithms.
\newblock In \emph{IEEE 11th Int Conf on Communication Software and Networks
  (ICCSN)}, pages 682--687, 2019.
\newblock \doi{10.1109/ICCSN.2019.8905362}.

\bibitem[Zychlinski(2018)]{zychlinski-tds}
Shaked Zychlinski.
\newblock The search for categorical correlation, 2018.
\newblock URL \url{https://towardsdatascience.com/}.

\bibitem[Stekhoven and Bühlmann(2011)]{stekhoven-missforest}
Daniel~J. Stekhoven and Peter Bühlmann.
\newblock {MissForest—non-parametric missing value imputation for mixed-type
  data}.
\newblock \emph{Bioinformatics}, 28\penalty0 (1):\penalty0 112--118, 10 2011.
\newblock ISSN 1367-4803.
\newblock \doi{10.1093/bioinformatics/btr597}.

\bibitem[Lin et~al.(2020)Lin, Chen, and Qi]{lin-deep19}
Enlu Lin, Qiong Chen, and Xiaoming Qi.
\newblock Deep reinforcement learning for imbalanced classification.
\newblock \emph{Appl. Intell.}, 50\penalty0 (8):\penalty0 2488--2502, 2020.
\newblock \doi{10.1007/s10489-020-01637-z}.

\bibitem[Hasselt et~al.(2016)Hasselt, Guez, and Silver]{dql-dmind}
Hado~van Hasselt, Arthur Guez, and David Silver.
\newblock Deep reinforcement learning with double q-learning.
\newblock In \emph{Proceedings of the 30th AAAI Conf on Artificial
  Intelligence}, AAAI'16, page 2094–2100. AAAI Press, 2016.

\bibitem[Chawla et~al.(2002)Chawla, Bowyer, Hall, and Kegelmeyer]{smote}
Nitesh~V. Chawla, Kevin~W. Bowyer, Lawrence~O. Hall, and W.~Philip Kegelmeyer.
\newblock {SMOTE:} synthetic minority over-sampling technique.
\newblock \emph{J. Artif. Intell. Res.}, 16:\penalty0 321--357, 2002.
\newblock \doi{10.1613/jair.953}.

\end{thebibliography}
\end{document}